%% file: root.tex
\documentclass[letterpaper, 10 pt, conference]{ieeeconf}  

\IEEEoverridecommandlockouts                              

\input{packages}
\input{macros}

\def\compileforpublish{1}

\title{\LARGE \bf
	A Knowledge-based Approach for the Automatic Construction of Skill Graphs for Online Monitoring*
}

\author{Inga Jatzkowski$^{1}$, Till Menzel$^{1}$, Ansgar Bock$^{2}$, and 
		Markus Maurer$^{1}$
\thanks{*This research is accomplished within the project ``UNICAR\emph{agil}''~\cite{WoopenUNICARagildisruptivemodulararchitectures2018}~(FKZ 16EMO0285). We acknowledge the financial support for the project by the German Federal Ministry of Education and Research~(BMBF).}
\thanks{$^{1}$Inga Jatzkowski, Till Menzel, and Markus Maurer are with the Institute of Control Engineering at TU Braunschweig, 38106 Braunschweig, Germany
        {\tt\small \{lastname\}@ifr.ing.tu-bs.de}} 
\thanks{$^{2}$Ansgar Bock wrote his master thesis at the Institute of Control Engineering, TU Braunschweig, 38106 Braunschweig, Germany}%
}

\begin{document}

\maketitle
\thispagestyle{empty}
\pagestyle{empty}

\input{sections/abstract}

\copyrightnotice

\section{INTRODUCTION}\label{sec:introduction}
\input{sections/introduction}

\section{SKILL GRAPHS}\label{sec:skillgraphs}
\input{sections/skillgraphs}

\section{ONTOLOGIES}\label{sec:ontologies}
\input{sections/ontologies}

\section{RELATED WORK}\label{sec:related_work}
\input{sections/stateoftheart}

\section{APPROACH}\label{sec:approach}
\input{sections/approach}

\section{PRELIMINARY RESULTS AND DISCUSSION}\label{sec:results}
\input{sections/results}

\section{CONCLUSION \& FUTURE WORK}\label{sec:conclusion}
\input{sections/conclusion}

\section{ACKNOWLEDGMENT}\label{sec:acknowledgment}
\input{sections/acknowledgment}

\renewcommand*{\bibfont}{\footnotesize} 
\printbibliography

\end{document}

%% file: packages.tex
\usepackage{color}
\usepackage{tubscolors}
\usepackage{tikz}
\usetikzlibrary{shapes,
				automata,
				arrows,
				arrows.meta,
				matrix,
				backgrounds,
				fit,
				patterns,
				decorations.markings, 
				svg.path, 
				shapes.multipart, 
				shapes.geometric, 
				external, 
				shadows, 
				patterns,
				positioning, 
				calc, 
				decorations, 
				scopes,
				fadings,
				bending,
				scopes}
\usepackage[simplified]{pgf-umlcd}
\usepackage{siunitx}
\usepackage[
			bibstyle=ieee,
			citestyle=numeric-comp,
			backend=biber,
			minnames=1,
			maxcitenames=2,
			maxbibnames=6,
			isbn=false,
			url=false,
			natbib=true,
			clearlang=true,
			date=year
			]{biblatex} 
\AtEveryBibitem{\clearlist{language}\clearfield{note}\clearfield{doi}}

\DeclareSourcemap{
	\maps[datatype=biber]{
		\map{
			\step[fieldsource=note, final]
			\step[fieldset=addendum, origfieldval, final]
			\step[fieldset=note, null]
		}
	}
}
\usepackage{xpatch}
\xpatchbibdriver{online}
{\printtext[parens]{\usebibmacro{date}}}
{\iffieldundef{year}
	{}
	{\printtext[parens]{\usebibmacro{date}}}}
{}
{\typeout{There was an error patching biblatex-ieee (specifically, ieee.bbx's @online driver)}}

\usepackage{ifthen}
\usepackage{lipsum}
\usepackage{amsmath}
\usepackage{amssymb} 	
\usepackage{bm} 		
\usepackage{booktabs} 	
\usepackage{graphicx}
\usepackage{microtype}
\usepackage{fmtcount}
\usepackage[hidelinks]{hyperref}



%% file: macros.tex
\makeatletter
\newcounter{IEEE@bibentries}
\renewcommand\IEEEtriggeratref[1]{%
	\renewbibmacro{finentry}{%
		\stepcounter{IEEE@bibentries}%
		\ifthenelse{\equal{\value{IEEE@bibentries}}{#1}}
		{\finentry\@IEEEtriggercmd}
		{\finentry}%
	}%
}
\makeatother

\newcommand\copyrighttext{%
	\footnotesize \parbox[t]{.11\textwidth}{\copyright{} \the\year~IEEE.} \parbox[t]{.89\textwidth}{Personal use of this material is permitted. Permission from IEEE must be obtained for all other uses, in any current or future media, including reprinting/republishing this material for advertising or promotional purposes, creating new collective works, for resale or redistribution to servers or lists, or reuse of any copyrighted component of this work in other works.}}
\newcommand\copyrightnotice{%
	\ifx \compileforpublish \undefined
	\else
	\begin{tikzpicture}[remember picture,overlay]
	\node[anchor=south,yshift=10.5pt] at (current page.south) {\parbox{\dimexpr\textwidth-\fboxsep-\fboxrule\relax}{\copyrighttext}};
	\end{tikzpicture}%
	\fi
}

%% file: sections/abstract.tex
\begin{abstract}
Automated vehicles need to be aware of the capabilities they currently possess.
Skill graphs are directed acylic graphs in which a vehicle's capabilities and the dependencies between these capabilities are modeled.
The skills a vehicle requires depend on the behaviors the vehicle has to perform and the operational design domain (ODD) of the vehicle.
Skill graphs were originally proposed for online monitoring of the current capabilities of an automated vehicle.
They have also been shown to be useful during other parts of the development process, e.g. system design, system verification.
Skill graph construction is an iterative, expert-based, manual process with little to no guidelines. 
This process is, thus, prone to errors and inconsistencies especially regarding the propagation of changes in the vehicle's intended ODD into the skill graphs.
In order to circumnavigate this problem, we propose to formalize expert knowledge regarding skill graph construction into a knowledge base and automate the construction process.
Thus, all changes in the vehicle's ODD are reflected in the skill graphs automatically leading to a reduction in inconsistencies and errors in the constructed skill graphs.
\end{abstract}

%% file: sections/introduction.tex
In order for an automated vehicle to operate safely in its environment, it must have knowledge of its current capabilities and whether they suffice for safe operation \cite{NolteSupportingSafeDecision2020}.
Skill and ability graphs have been proposed as a framework for modeling and monitoring of the (current) capabilities of automated vehicles \cite{ReschkaAbilityskillgraphs2015}. 
The construction of these graphs is done manually by experts who possess a thorough understanding of the system and the intended operational design domain (ODD) \cite{SAEJ3016TaxonomyDefinitions2018}.
This construction process is an ad-hoc process following no clear directions or guidelines leaving the experts without a clear starting point or idea of when a graph is complete.

Skill graphs as proposed in \cite{ReschkaAbilityskillgraphs2015} are constructed as a directed acyclic graph of the skills necessary to perform an abstract behavior, e.g. a driving maneuver, and the dependencies between these skills.
As several behaviors can require the same skills, these graphs may partially overlap.
Manual construction of the skill graphs for an automated vehicle, as any manual modeling process, is error prone.
Practical experience has shown that the experts constructing the graphs may forget crucial skills or dependencies during the modeling process. 
Skill graphs are designed iteratively and adjusted during the development process, thus changes in the graph have to be tracked especially for overlapping parts of the graphs to prevent inconsistencies.
Even when the initial graphs were consistent and were constructed correctly, integration and tracking of changes in the graphs proves to be a challenge for human modelers.

Experts are usually an expensive resource.
Rather than having experts perform the entire modeling task including checking for inconsistencies between the graphs for the individual behaviors, it would be more efficient to automate as much of the skill graph construction process as possible.
Thus, formalizing the experts knowledge as well as the construction process itself can reduce expert involvement in the modeling process.
Experts can be more effectively utilized to produce reusable artifacts for the modeling process and to evaluate the result of the automated modeling process.

In previous works, the modeled capabilities and the intended ODD were either small \cite{MaurerFlexibleAutomatisierungStrassenfahrzeugen2000,SiedersbergerKomponentenzurautomatischen2003,PellkoferVerhaltensentscheidungfuerautonome2003} or the construction of skill graphs was only demonstrated for one or a few selected behaviors \cite{ReschkaAbilityskillgraphs2015,ReschkaFertigkeitenundFaehigkeitengraphen2017,Nolteskillabilitybaseddevelopment2017}.
The construction of a full set of skill graphs for a fully automated vehicle capable of performing a range of behaviors in a complex ODD has not been presented so far.
Thus, the challenges accompanying the construction of skill graphs for multiple behaviors in a complex ODD have not arisen before and a structured and formalized construction process was not necessary due to the reduced complexity of the task.


To handle the complexity of the construction process, we propose to design the construction process to require only minimal expert involvement. 
Thus, expert knowledge is composed into a knowledge base. 
Every vehicle behavior requires a foundation of skills for its execution. 
Modeling these foundation skills still involves experts with knowledge of the respective behaviors.
Additional necessary skills depend on the scene elements present in the vehicle's ODD.
These additional skills are inferred from the ODD and automatically added to the base graph of foundation skills using the information stored in the knowledge base.
Experts should be involved again in validating the generated graphs.

This process relieves experts from the tedious parts of skill graph construction while keeping them involved for the aspects where their expertise is indispensable.
Generating the ODD-dependent part of the graphs automatically has the additional advantage that changes in the ODD are directly reflected in the skill graphs.
At this point, only skill requirements derived from the ODD are reflected in this knowledge-based construction process.
It is conceivable that skills may also depend on other aspects such as traffic rules or Object and Event Detection and Response (OEDR) strategies.
However, additional dependencies can be easily added to the knowledge base.
Another possible advantage of a knowledge-based automatic generation of skill graphs is that it is sufficient to verify the correctness of the knowledge base instead of the correctness of every graph.
Correctness of the graphs is guaranteed due to the reasoning of the ontology as long as the information inside the knowledge base is correct and complete.

The remainder of this paper is structured as follows:
\autoref{sec:skillgraphs} and \autoref{sec:ontologies} give a brief overview of the concept of skill graphs and the concept of ontologies for knowledge representation.
In \autoref{sec:related_work}, we provide an overview of relevant related publications before we present our approach for automatic skill graph construction and illustrate it with an example in \autoref{sec:approach}.
We discuss preliminary results and limitations of this approach in \autoref{sec:results} and conclude the paper in \autoref{sec:conclusion}. 

%% file: sections/skillgraphs.tex

\begin{figure}[htb]
	\centering
	\includegraphics[width=0.94\linewidth]{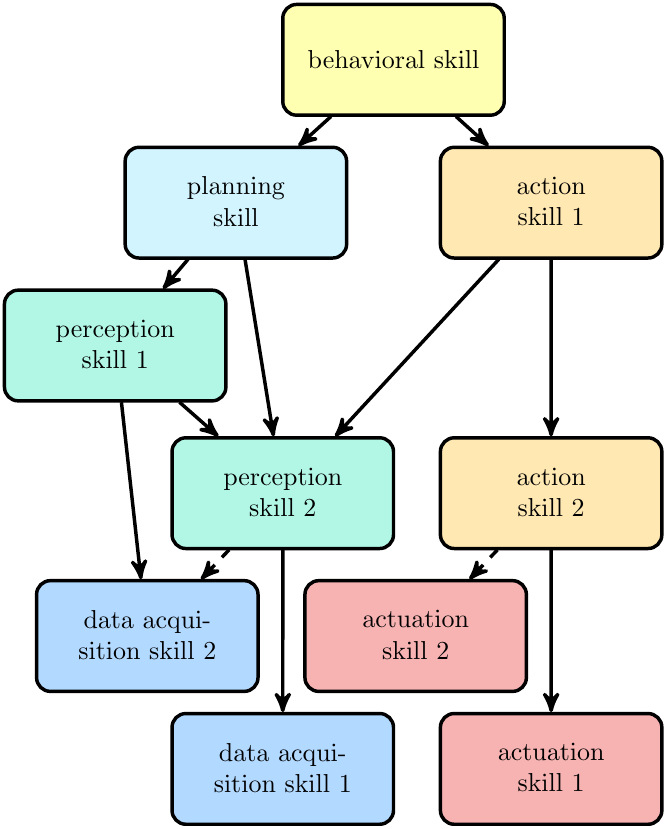}
	\caption{Example of the structure of a skill graph. 
		Boxes represent skill nodes, colors denote the skill categories: behavioral (yellow), action (orange), actuation (red), planning (light blue), perception (green), data acquisition (dark blue).
		Solid arrows represent 'depends on'-relations, dashed arrows represent 'may depend on'-relations.}\label{fig:example_graph}
	\vspace{-0.5cm}
\end{figure}

Skill graphs were introduced by \citet{ReschkaAbilityskillgraphs2015} and are based on the concept of a skill network presented in \cite{MaurerFlexibleAutomatisierungStrassenfahrzeugen2000,SiedersbergerKomponentenzurautomatischen2003,PellkoferVerhaltensentscheidungfuerautonome2003,BergmillerFunctionalSafetyDrivebyWire2015}.
Skill graphs are directed acyclic graphs.
The nodes of the graph represent skills and the directed edges between the nodes represent "depends on" relations between the skills.
The level of abstraction within the skill graph is highest at the root of the graph and becomes less abstract towards the leaves.
Each skill in the skill graph belongs to one of the following seven categories: system skills, behavioral skills, planning skills, perception skills, data acquisition skills, action skills, and actuation skills.
In earlier publications \cite{ReschkaAbilityskillgraphs2015,ReschkaFertigkeitenundFaehigkeitengraphen2017,Nolteskillabilitybaseddevelopment2017,KnuppelSkillBasedVerificationCyberPhysical2020}, data acquisition skills and actuation skills were titled data sources and data sinks respectively.
However, data sinks and data sources are objectively not skills the same way that eyes or legs are not skills \cite{NoltePersonalcommunication2020}.
The underlying skills are the acquisition of sensory data (from sensor hardware or the optic nerve) and the control of the actuators (controlling actuator hardware or the capability to move the legs).
We, therefore, adjust the terminology accordingly.
The aforementioned skill categories form a hierarchy based on their level of abstraction, meaning a skill of one category can only have child nodes of specific other categories, c.f. \cite{KnuppelSkillBasedVerificationCyberPhysical2020}.
Data acquisition skills and actuation skills form the leaves of the graph and have no child nodes.
An example graph showing the general graph structure is depicted in \autoref{fig:example_graph}.

%% file: sections/ontologies.tex

According to \citeauthor{NicolaGuarinoWhatOntology2009}, "an ontology is a formal, explicit specification of a shared conceptualization" \cite[p. 2]{NicolaGuarinoWhatOntology2009}.
A conceptualization formally represents the entities that are of interest and the relationships that hold among these entities for a domain of interest.
'Formal' refers to the fact, that the representation must be machine-readable.
Ontologies should also be human-readable as they facilitate the communication between human and machine.
A human- and machine-readable formal representation is achieved by using a subset of first-order predicate logic reduced to unary and binary predicates as a language for representing knowledge.
Concepts also called classes are described by unary predicates, roles also called relations or properties are described by binary predicates, and individuals are instances of a concept or class \cite{NicolaGuarinoWhatOntology2009}.

Ontologies are structured into terminological boxes (T-box) describing the concepts of a domain, i.e. hierarchical classes, axioms, and properties, and assertional boxes (A-box) representing individuals of classes and knowledge from data.

Reasoners can infer additional knowledge from terminological and assertional boxes, identify conflicts in concept and axiom definitions, and check for consistency \cite{HulsenTrafficintersectionsituation2011b}.

%% file: sections/stateoftheart.tex

Skill graphs were proposed for online capability monitoring in \cite{ReschkaAbilityskillgraphs2015} and further substantiated in \cite{ReschkaFertigkeitenundFaehigkeitengraphen2017,Nolteskillabilitybaseddevelopment2017}.
Skill graphs model the skills necessary for a vehicle to perform an abstract behavior as nodes in a directed acyclic graph and the dependencies between these skills as directed edges between nodes.
\citet{ReschkaFertigkeitenundFaehigkeitengraphen2017} also proposes the use of skill graphs during the development process to aid in the construction of a functional system architecture.
\citet{Nolteskillabilitybaseddevelopment2017} extend the use of skill graphs in early stages of the development process by demonstrating their usefulness for the derivation and refinement of functional requirements from safety-requirements along the skills in a skill graph.
\citet{BagschikSystemPerspectiveArchitecture2018} propose to regard skill graphs as one view in an architecture framework that is connected to other architecture views such as the software, hardware, or functional system architecture.
Skill graphs provide a functional viewpoint independent from the implementation realized in software or hardware and independent from the representation of functional components and interfaces.
Through interconnections with other architecture views, skills can be related to their implementation or functional system components.
\citet{KnuppelSkillBasedVerificationCyberPhysical2020} utilize skill graphs for the verification of cyber-physical systems. 
The authors combine skill graphs as a formal system model with a formal theorem prover.
They connect the individual skills of a skill graph with models for the realization of the skills and show a verification of a skill graph regarding safety requirements for the skills.
\citeauthor{KnuppelSkillBasedVerificationCyberPhysical2020} also provide a formalization of skill graphs. 
While several possible applications for skill graphs have been proposed, none of the publications provide a structured process for the construction of skill graphs.

\citet{ColwellAutomatedVehicleSafety2018} note that a change in capabilities of the vehicle results in a restriction of the ODD the vehicle is able to operate in safely. 
They define one or more so-called degraded operation modes caused by system impairments for each subsystem of the automated vehicle and relate these modes to restrictions of the ODD.
While \cite{ColwellAutomatedVehicleSafety2018} do not make use of the skill graph concept to manage ODD restrictions, they note that skill graphs could provide a useful abstraction between degraded operation modes and ODD restrictions. 
This connection of ODD and required vehicle skills is also stated in \cite{NolteSupportingSafeDecision2020}.
\citet{NolteSupportingSafeDecision2020} provide a taxonomy of self-monitoring concept for automated vehicles and relate skill graphs as a capability representation to other aspects of self-representation and to the ODD.
They state that the ODD determines the necessary capabilities of an automated vehicle as well as the functional requirements these capabilities have to fulfill.

Several recent publications have focused on a formal description of the ODD \cite{BritishStandardsInstitutionPAS1883Operational2020,KoopmanHowManyOperational2019,GyllenhammarOperationalDesignDomain2020,CzarneckiOperationalWorldModel2018a, CzarneckiOperationalWorldModel2018}.
While they differ in the details, they all include a representation of scene elements.
For a scene representation, \citet{BagschikOntologybasedScene2018} propose a five-layer model to structure scene elements such as traffic participants and their interactions, and environmental influences.
They demonstrate the usefulness of this model in a knowledge-based scene generation for German highways.
While not initially intended for ODD description, the five-layer model can be utilized to structure scene elements in an ODD description.
Similar approaches for a representation of (parts of) the ODD in a knowledge base were presented in \cite{HummelSceneunderstandingurban2008} and \cite{HulsenTrafficintersectionsituation2011b}.

%% file: sections/approach.tex

\begin{figure}[htb]
	\centering
		\includegraphics[width=\linewidth]{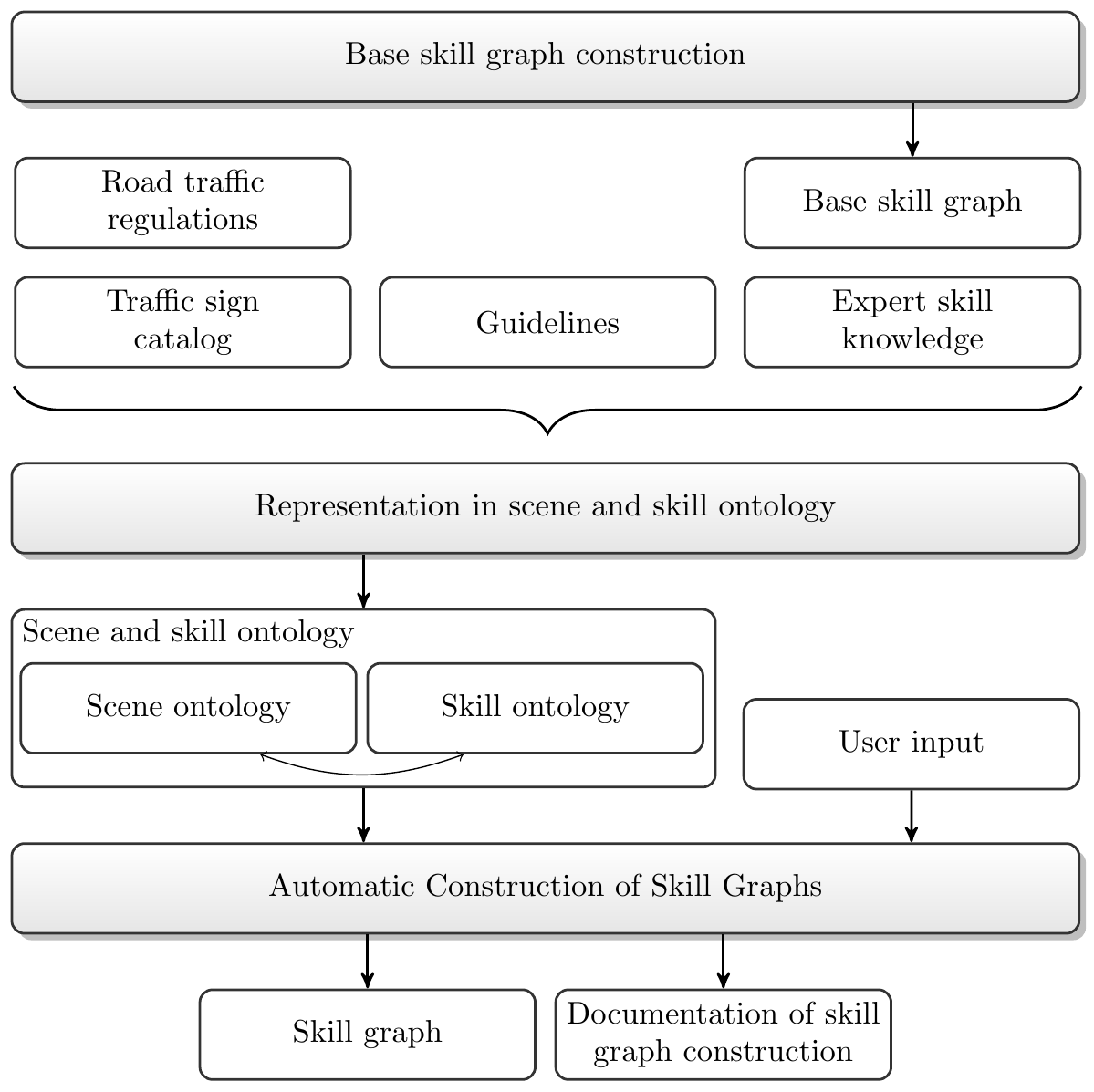}
	\caption{Overview of the process for automatic skill graph construction. Gray boxes represent process steps, white boxes represent inputs/outputs.}\label{fig:process}
\end{figure}

To enable automatic skill graph construction, several steps of manual information processing are required beforehand.
An overview of the process is shown in \autoref{fig:process}.
In a first step, base skill graphs must be constructed by experts for skills and behaviors.
The base skill graphs, along with expert knowledge about skills, and regulatory information concerning scene elements are represented in a scene and skill ontology.
This ontology and a user's selection of scene elements form the input for a python-based implementation realizing the automatic skill graph construction according to stored rules.

Requirements for the necessary skills of an automated vehicle stem from at least two sources: the behaviors the vehicle shall be able to perform and the ODD the vehicle shall perform these behaviors in.
Every behavior requires a set of foundation skills to perform it regardless of the intended domain.
Thus, a base skill graph can be constructed from these foundation skills for every behavior. 
The construction of these base graphs is a task for experts as it requires deeper knowledge about what each behavior entails. 
However, the construction of the base skill graphs only has to be done once.
The base skill graphs for the behaviors are ODD-independent and can be reused for different domains as long as the required behaviors do not change.
\begin{figure*}[bt]
	\centering
	\includegraphics[width=\textwidth]{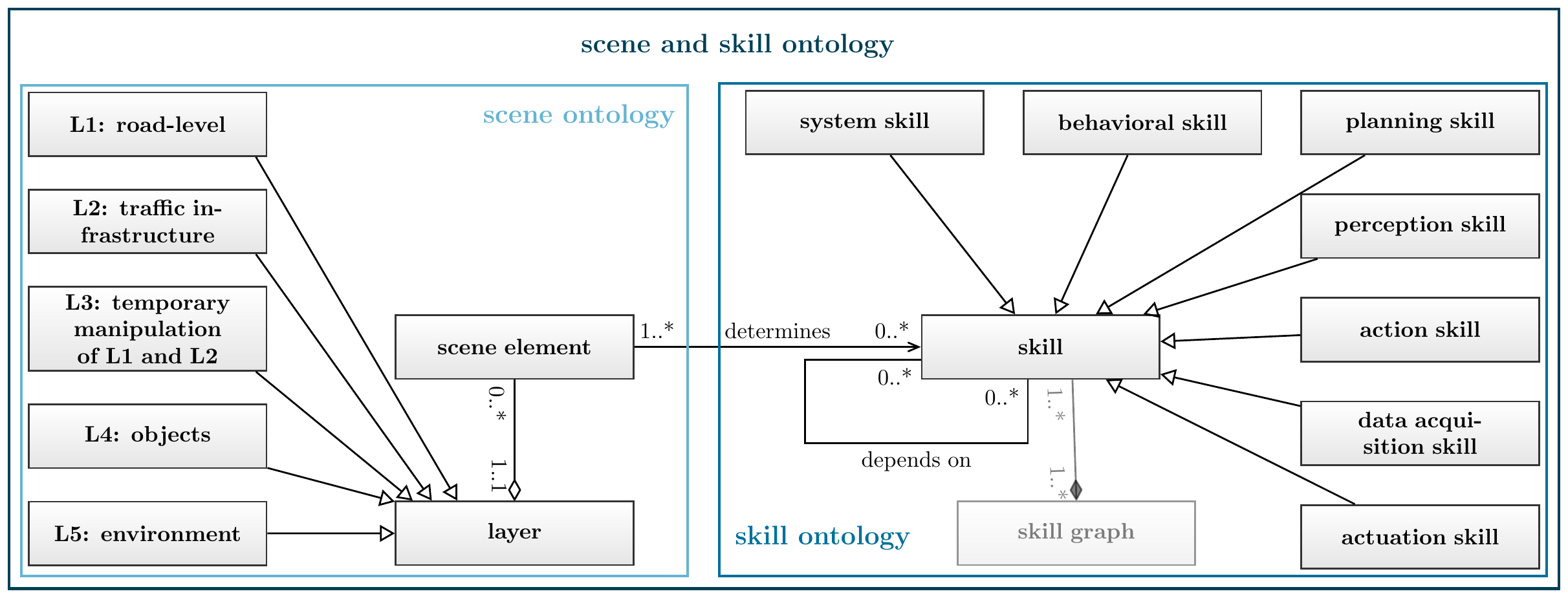}
	\caption{Class diagram of the connections between skills and scene elements.}\label{fig:class-diagramm}
	\vspace{-0.5cm}
\end{figure*}
Once the base skill graphs have been constructed, requirements for additional skills can be derived from the ODD.
The ODD plays a central part in the development of an automated vehicle and several approaches for its description have been proposed \cite{BritishStandardsInstitutionPAS1883Operational2020,KoopmanHowManyOperational2019,GyllenhammarOperationalDesignDomain2020}.
What they all have in common is that they describe the scene elements \cite{UlbrichDefiningSubstantiatingTerms2015} that can occur in the ODD.
To structure these scene elements, the five-layer model for the representation of driving scenes by \citet{BagschikOntologybasedScene2018} can be used.
It structures the scene elements in the following five layers:
\begin{itemize}[\setlabelwidth{LX}]
	\item[L1:] road-level elements
	\item[L2:] traffic infrastructure
	\item[L3:] temporary manipulation of L1 and L2
	\item[L4:] objects
	\item[L5:] environmental conditions
\end{itemize}
Knowledge regarding the scene elements of these layers can be modeled in an ontology as demonstrated in \cite{BagschikOntologybasedScene2018} for German highways.
By extending the ontology in \cite{BagschikOntologybasedScene2018} with the base skill graphs for the behaviors, the skill(s) necessary for handling a scene element, and dependencies between skills, this extended scene and skill ontology can be used to automatically generate a skill graph for a certain behavior and a specific ODD.
The structure of the scene and skill ontology is depicted in \autoref{fig:class-diagramm}.

In order to use this scene and skill ontology for automated skill graph construction, it is necessary to access the information stored within the ontology and infer additional information from the stored properties.
A python-based implementation with a QT-based graphical user interface was programmed to access the information stored in the ontology and utilize it for automated skill graph construction. 
The implementation utilizes the python library Owlready2~\cite{LamyOwlreadyOntologyorientedprogramming2017} to access the information in the ontology.
Via the GUI, a behavior and a general domain, e.g. highway, can be selected. 
The domain can be further specified by manipulating the occurring scene elements.
This input is used by the underlying implementation to access the ontology and infer the skill graph for the specified behavior and the selected ODD.

In the following, we describe the individual steps of the approach in detail, i.e. the construction of the base skill graphs, building the scene and skill ontology, and the automatic skill graph construction.
Each step is illustrated using the behavior "lane keeping" as an example.

\subsection{Construction of base skill graph}
The base skill graph is constructed by experts with knowledge of the behavior the vehicle shall perform.
Every vehicle behavior is connected to some basic infrastructure that needs to be present for the vehicle to be able to perform the behavior.
Every behavior requires a driving surface and individual infrastructure elements.

This selection of minimum necessary scene elements can aid experts in the derivation of the foundation set of skills a vehicle requires to perform the behavior.
These skills and the dependencies between these skills form the base skill graph for the behavior and are always required regardless of the ODD.
This approach was inspired by the utilization of a base case in a maneuver description in \cite{CzarneckiAutomatedDrivingSystem2018}.

\subsubsection{Example: Lane keeping}
\begin{figure}[htb]
	\centering
	\includegraphics[width=0.68\linewidth]{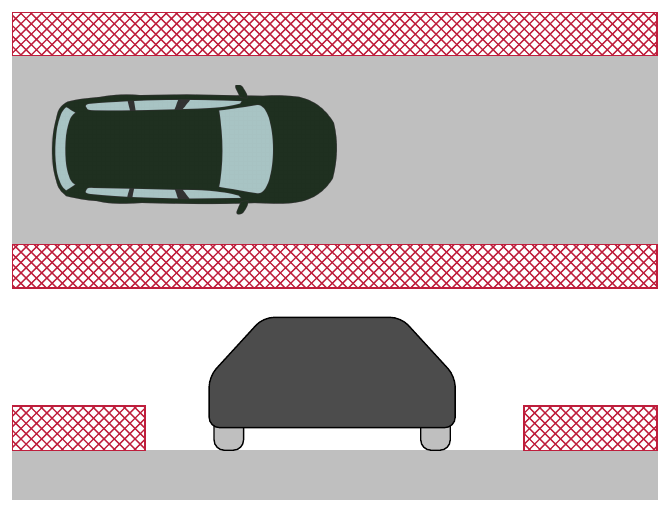}
	\caption{Visual representation of the selection of scene elements which must be present for a vehicle to perform the behavior "lane keeping". The gray area depicts the road surface or surface of the drivable area. The areas with red crosshatch depict unspecified lane boundaries.}\label{fig:base_scene}
	\vspace{-0.5cm}
\end{figure}

We will illustrate the base skill graph construction at the example of the behavior "lane keeping".
This behavior comprises the lateral aspects of following a lane but not the longitudinal aspects.
The behavior lane keeping requires the existence of at least one lane on a drivable area with some form of lane boundaries.
The lane boundaries are intentionally kept vague as it is only relevant that there is some way of discerning where the lane ends but not how.
At this point the generalized unspecific concept of lane boundaries includes all possible variations in the field: lane markings, implicit boundaries of the drivable area, curbs, virtual boundaries stored in a digital map.
A visual representation of the selection of minimum necessary scene elements is depicted in a graphical format in \autoref{fig:base_scene}.

\begin{figure}[tb]
	\centering
	\includegraphics[width=\linewidth]{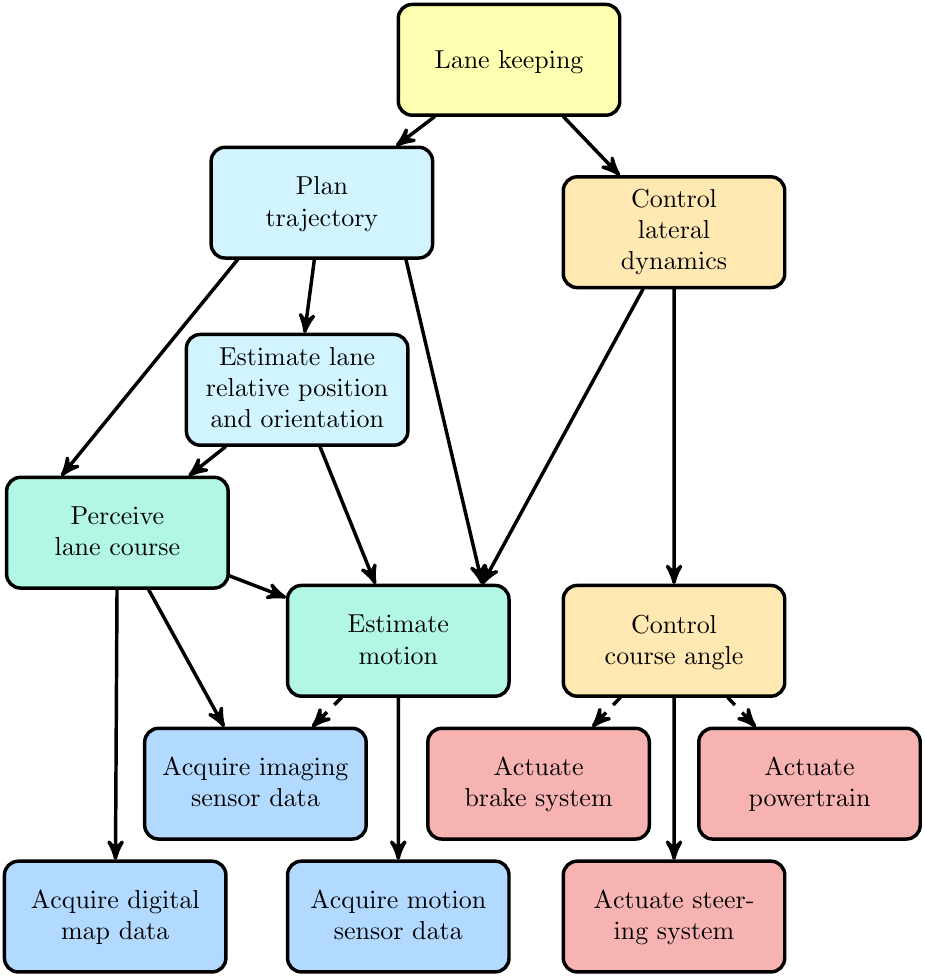}
	\caption{	Base skill graph for 'lane keeping'. 
				Notation see \autoref{fig:example_graph}}\label{fig:base-graph}
			\vspace{-0.5cm}
\end{figure}

The resulting base skill graph is depicted in \autoref{fig:base-graph}.
To follow a lane, the vehicle needs to be able to plan its trajectory to stay within the lane boundaries.
Thus, it needs to be able to perceive the course of the lane, estimate its position and orientation relative to the lane, and estimate its own vehicle motion.
The perception of the lane course requires either the evaluation of acquired digital map data and a pose estimate, or the evaluation of acquired imaging sensor data and a pose estimate.
Vehicle motion may be estimated by the evaluation of acquired motion sensor data or the evaluation of acquired imaging sensor data.

Additional to planning and perception skills, lane keeping also requires action skills.
To stay within lane boundaries, the vehicle must be able to control its lateral motion.
Thus, it needs to be able to control the course angle of the vehicle and it needs an estimate of the vehicle's motion.
This requires the skill of controlling the steering system.
It may also be realized by controlling the powertrain or the brake system.

Skills closer to the root are more abstract and are necessary for most behaviors.
Skills closer to the leaves are more specific and depend more on the ODD.
The actuation skills at the leaves are fixed due to the general actuator design of a vehicle.
The data acquisition skills are intentionally kept vague and are only separated into evaluation of digital map data, evaluation of imaging sensor data, and evaluation of motion sensor data. 
In this way, skill graphs can assist in deriving a sensor concept based on the required skills.

\subsection{Scene and skill ontology}

The scene and skill ontology contains the connections between scene elements and required skills, the connections between individual skills, and connection between behaviors and the base skill graphs.
An ontology provides a format to organize information, i.e. data and their semantic connections, in a human- and machine-readable manner, c.f. \autoref{sec:ontologies}.
The scene and skill ontology is a simplification and an extension of the scene ontology in \cite{BagschikOntologybasedScene2018}.

The scene elements in the scene and skill ontology are structured using the five-layer-model for the representation of driving scenes \cite{BagschikOntologybasedScene2018}.
Scene elements for the domains (German) highways and urban areas were included in the ontology.
The scene elements were derived from guideline documents for the construction of highways \cite{ForschungsgesellschaftfurStrassenundVerkehrswesenRichtliniefurAnlage2008} and of urban roads \cite{ForschungsgesellschaftfurStrassenundVerkehrswesenRichtliniefurAnlage2007}, and from German traffic regulations \cite{BundesministeriumfurVerkehrStrassenverkehrsordnungStVO2013}.
Each scene element belongs to none or multiple domains but belongs to exactly one layer. 

Skills are structured using the seven skill categories introduced in \autoref{sec:skillgraphs}.
Each skill belongs to exactly one skill category.
The skills were derived using expert knowledge as no skill catalog exists for all the skills an automated vehicle requires.
Dependency relations between the individual skills were also derived from expert knowledge and added to the skills as properties.
A skill can depend on none or multiple other skills.
Only actuation and data acquisition skills depend on no other skills.
Skills are connected via a dependency relation representing the edges between skills in the skill graph.
For each behavioral skill, all skills forming the base skill graph were added using a separate necessity relation.
The same necessity relation is used to model relationships between skills that exclusively occur together in a skill graph.

Finally, the scene element part and the skill part of the ontology were connected via relations between scene elements and skills.
A scene element can determine the necessity for none or multiple skills.
A skill can also be determined by multiple scene elements.
This relation is modeled as a property of the individual scene elements but could just as well be modeled as a skill property.

For skill graph construction, it is only relevant whether a scene element exists within the ODD.
The placement of elements is (mostly) irrelevant. 
Thus, connections between scene elements necessary for automatic scene creation as in \cite{BagschikOntologybasedScene2018} were not included in the ontology.
For simplicity, temporary manipulations of road-level elements and traffic infrastructure (L3), such as road works, were omitted from the ontology as well.
Layer 4, objects, includes the interactions between objects and the maneuvers dynamic objects perform. 
Maneuvers are a representation of behaviors for which the skill graphs are constructed, and, thus, are connected to behavioral skills as the root nodes of the individual skills graphs.
The environmental conditions on layer 5 may influence the quality of a skill but do not evoke requirements for a skill's existence and were, thus, not used for skill graph construction.
In essence, only scene elements of layers~1,~2, and layer~4 were used for the skill graph generation.

The resulting ontology contains the experts knowledge about the scene elements that can be present in a domain, the skills determined due to the presence of these scene elements, and the dependencies between all skills.

\subsection{Automated skill-graph construction}
In order to use the scene and skill ontology for automated skill graph construction, we need to be able to access the information stored in it and infer additional information from the stored properties.
The information stored in the ontology in the form of classes and properties represent the T-box.
As stated above, a python-based implementation with a graphical user interface (GUI) was programmed to access the information stored in the ontology and utilize it for automated skill graph construction. 
Via the GUI, a behavior and a general domain, e.g. highway domain or urban domain, can be selected. 
For the selected domain, the occurring scene elements can be further specified.
The selection of behavior and scene elements generates instances of the respective classes stored in the T-box.
These instances are added to the A-box and are used as input for the underlying implementation to access the information in the T-box and infer the skill graph for the selected behavior and the specified ODD.

During the inference process the A-box is populated with instances of skills inferred from the existence of scene elements and from necessity relations between skills.
The behavioral skill determined by the behavior scene element is extracted from the ontology and an instance of the class is added to the A-box.
At the start of the inference process, the base graph for the behavior is extracted from the ontology as the behavioral skill has direct necessity relations to these skills of the base skill graph.
Instances of the skills of the base skill graph and the dependencies between these instances are added to the A-box.
As the base skills have dependency relations between each other, the base skill graph can be constructed according to these relations.
Then, the selected scene elements are used to infer additional skills using the association of scene elements to required skills.
Instances of these skills and their dependencies to other skills are added to the A-box.
Does an added skill necessitate a skill that is not part of the A-box yet an instance of the missing skill is added as well.
This is a recursive process until all missing skills are added.
Once all scene elements have been considered and all resulting instances of skills and their dependencies have been added to the A-box, the inference process is complete.
The resulting skill graph is extracted from the A-box and transformed into a suitable output format.
The implementation outputs the final graph and, additionally, a document detailing the insertion process for traceability of the modeling steps.

It may be worth noting at this point that it is not strictly necessary for an automated vehicle to possess all skills included in the generated skill graph.
Some requirements from the existence of certain scene elements may have different redundant solutions requiring different underlying skills.
All redundant solutions are modeled in this initial generation.
This allows the approach presented here to be used from the start of the development process.
\citet{ReschkaFertigkeitenundFaehigkeitengraphen2017} proposes skill graphs as a tool to aid in the creation of the functional system architecture.
We would also suggest the use of skill graphs for modeling possible redundant solutions to a problem and to help guide system implementation.
A highly detailed graph is likely not useful for capability monitoring considering the necessity of very detailed monitoring metrics that may be difficult to provide.
Therefore, we suggest to generate a very detailed initial skill graph for each behavior and prune and condense it as needed later in the development process to make it suitable for other application.
This pruning process may be automated as well.

\subsubsection{Example: Lane keeping}
\begin{figure}[tb]
	\centering
	\includegraphics[width=1.0\linewidth]{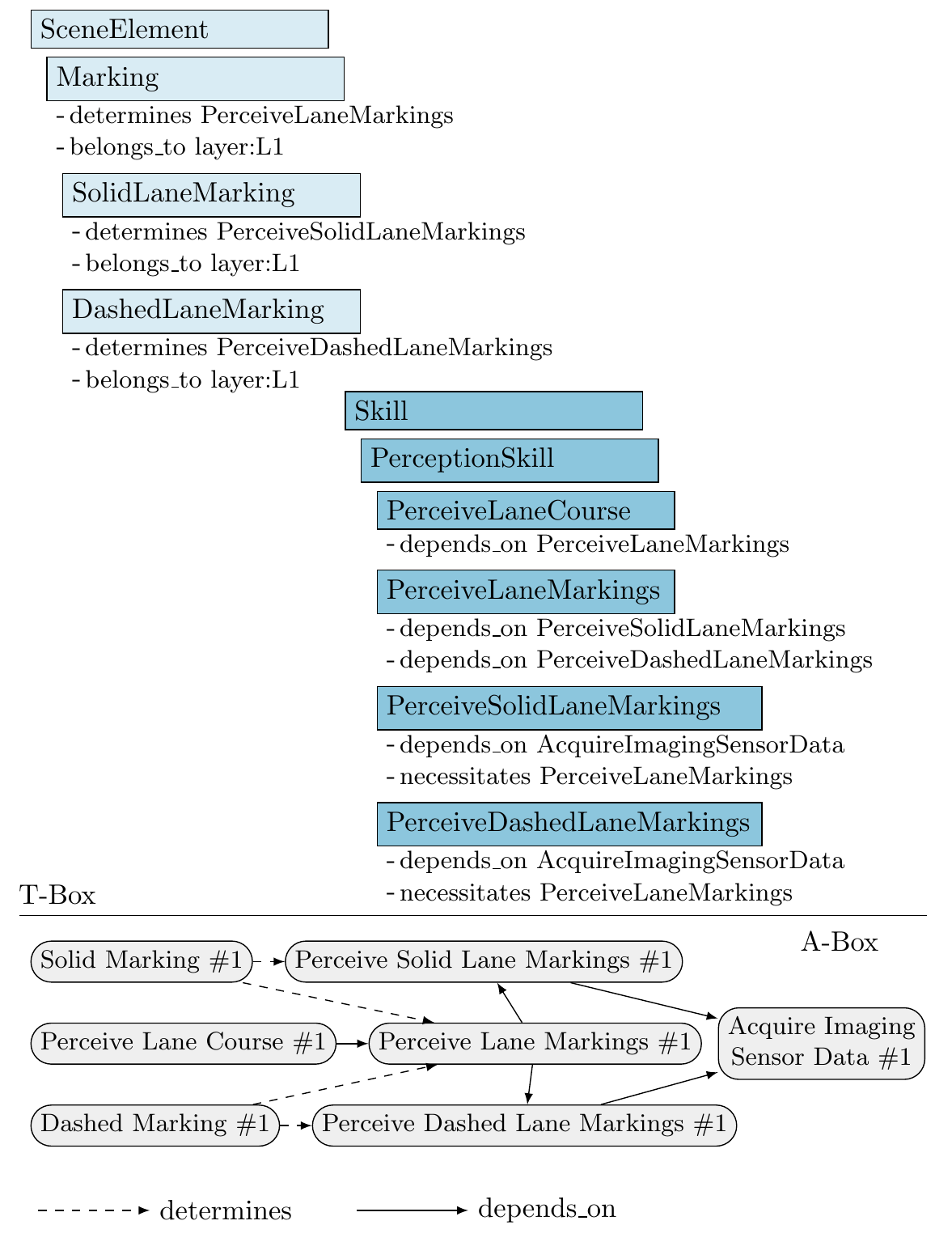}
	\caption{Excerpt of T-box and A-box for 'lane keeping' example with scene elements, solid marking and dashed marking, and related skills.}\label{fig:t-box_a-box}
	\vspace{-0.5cm}
\end{figure}

\begin{figure}[tb]
	\centering
	\includegraphics[width=\linewidth]{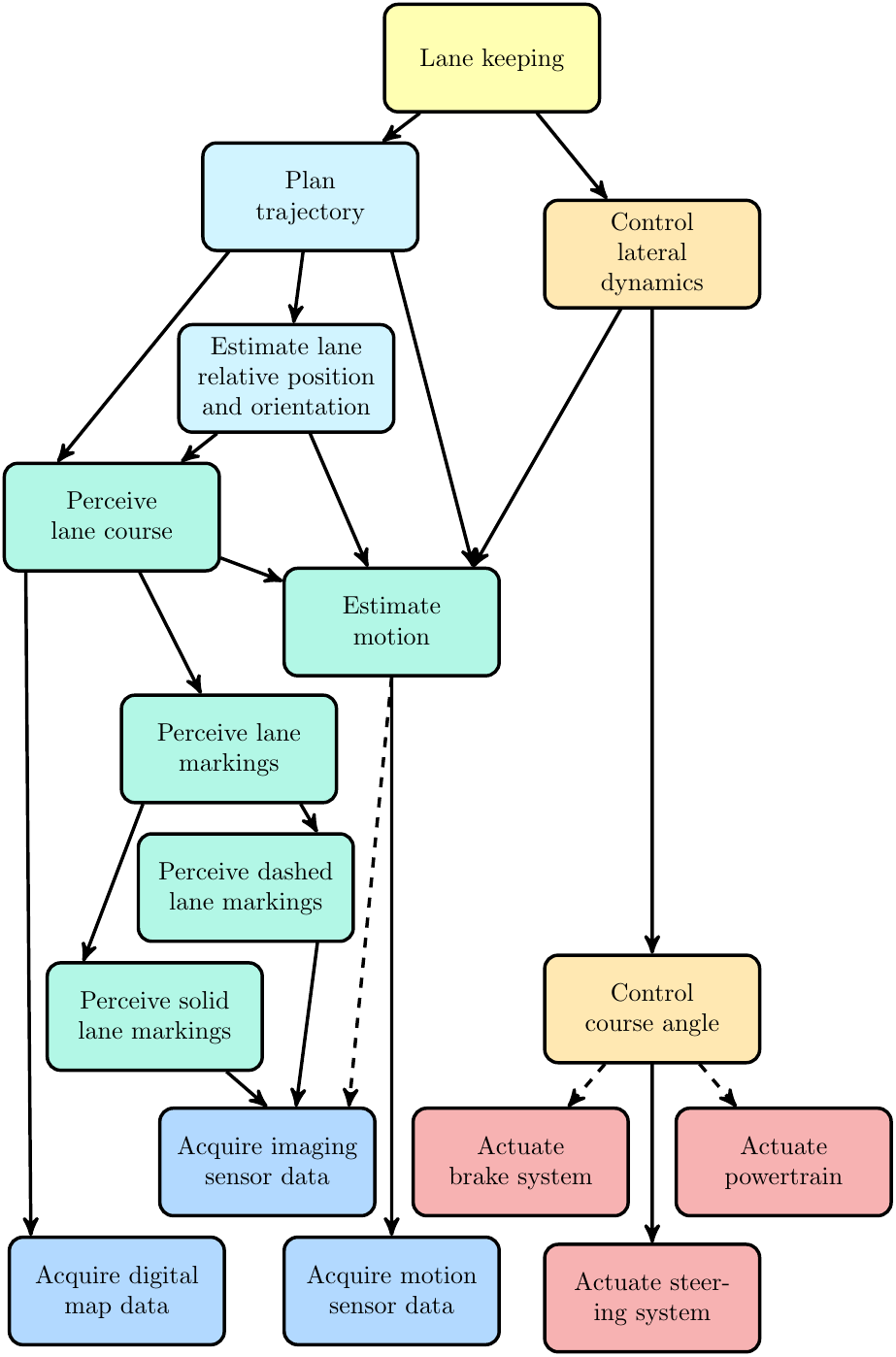}
	\caption{Skill graph for 'lane keeping' for an ODD with the scene elements 'solid lane marking' and 'dashed lane marking' on L1.
				Notation see \autoref{fig:example_graph}}\label{fig:graph1}
			\vspace{-0.5cm}
\end{figure}

We will illustrate the process of skill graph generation using the example of the "lane keeping" behavior discussed above.
\autoref{fig:t-box_a-box} shows part of the T-box for this example to illustrate the inference process.
Properties of more abstract classes are inherited by their child classes.
We select the behavior 'lane keeping', the domain, e.g. 'urban', and solid lane markings and dashed lane markings as explicit delineation between lanes on layer~1 as scene elements present in our ODD via the GUI.
If the ODD also contains roads with multiple lanes not delineated by lane marking this must be modeled as well, as the absence of lane markings may require additional skills.
This also applies to other infrastructure elements such as, e.g. stop lines at intersections.
Instances of the scene elements (behavior and lane markings) are added to the A-box.
The scene element 'lane keeping' is connected to the behavioral skill 'lane keeping' via a relation stored in the T-box, therefore an instance of the skill 'lane keeping' is added to the A-Box.
This behavioral skill necessitates the skills of the base skill graph which are extracted from the ontology and instances of the skills and their dependency relations are added to the A-box. 
The existence of any type of (lane) marking within the ODD determines the skill to 'perceive lane markings'. 
This property is inherited by the scene elements 'solid lane marking' and 'dashed lane marking' from the super-class 'marking'.
An instance of a determined skill is only added to the A-box once if it is determined by multiple scene elements.
The two different types of lane markings also determine the skills 'perceive solid lane markings' and 'perceive dashed lane markings'. 
Instances of these skills are added to the A-box according to the dependency relations stored in the skills' properties.
If in this example the skill 'perceive lane markings' were not determined by the existence of a scene element it would have been added to the A-box based on its necessity relations.
Both the skills 'perceive solid lane markings' and 'perceive dashed lane markings' necessitate the skill 'perceive lane markings'.
\autoref{fig:t-box_a-box} shows part of the A-box for this example.
The resulting skill graph extracted from the A-box is depicted in \autoref{fig:graph1}.
As stated above, it is not strictly necessary for an automated vehicle to perceive lane markings and infer the course of the lane from the perceived markings.
Extracting the course of the lane from the evaluation of digital map data and using a map-relative pose estimate of the vehicle is also a possible way to determine the lane relative position and orientation of the vehicle.
Both solutions are modeled in the graph and can be pruned later if desired.

%% file: sections/results.tex

Skill graphs for several different vehicle behaviors and a variety of scene element combinations were automatically generated using the approach presented above.
The base skill graphs for each of these behaviors were constructed manually by experts.
The generated skill graphs were analyzed by experts and found to be sound in their general construction.
No skills or dependencies were missing from the graph and all dependency connections were correctly drawn.

Automatically generated skill graphs still require expert assessment after generation to account for possible gaps in the knowledge base.
However, the automatic generation of skill graphs reduced errors in the construction process compared to a manual construction.
A number of automatically generated graphs were compared with manually constructed graphs.
This comparison managed to highlight inconsistencies in between the manually constructed graphs, missing dependency relations, and in rare cases missing skills.
Errors in the automatically generated graphs can be traced back to errors or gaps in the knowledge base using the automatically generated documentation of the skill graph construction steps.
Once the knowledge base is corrected, the errors in all affected graphs are corrected.
The construction process is, thus, mostly reduced to a review process.

This initial implementation serves mostly as a proof of concept and has several limitations.
In this initial implementation, the automatically generated skill graphs have a very fine skill granularity.
Meaning, for example, that every individual traffic sign type present in the ODD will require a skill to perceive this particular type of sign.
In order to derive requirements for system implementation, skill graphs with such a fine granularity can be helpful.
For purposes of capability monitoring during operation, such a fine granularity is most likely not useful.
For capability monitoring it is more relevant that traffic signs in general can still be perceived rather than every individual sign type.
Additionally, monitoring metrics for the perception quality of traffic sign detection in general can be more easily provided than the perception quality of each individual traffic sign type.
Thus, different levels of abstraction in skill granularity are necessary for different application.
One solution can be to define superordinate skills and group related skills under these super-skills.
Depending on a selected level of granularity only the super-skills are included in the final graph or the super-skills with all their sub-skills.
Adding granularity can increase the usefulness of the approach for different applications.

\citeauthor{BagschikOntologybasedScene2018} use the highway scene ontology presented in \cite{BagschikOntologybasedScene2018} to generate traffic scenes from the ontology.
During this process, they automatically exclude impossible combinations and relations of scene elements in the generated scenes.
While automated skill graph construction does not require the same semantic information as scene generation, semantic information could improve the selection of scene elements for the definition of the ODD.
ODD specification could be improved by including semantic information, e.g., about scene elements that are interdependent, meaning one element will never occur without the other, and therefore cannot be separated.
These connections can be unidirectional or bidirectional.
To include these semantic relations in the scene ontology can limit the mistakes made during ODD specification.

%% file: sections/conclusion.tex

In this paper, we proposed a knowledge-based approach for the automatic construction of skill graphs.
Automating the construction of skill graphs relieves experts of a tedious and error prone modeling task and allows to integrate changes in the graphs automatically.
Automating this construction process also means, that non-experts can generate skill graphs to use in other parts of the development process.
Experts will still be necessary to review the generated graphs.

We stated the influence of the ODD on the required capabilities of an automated vehicle in a previous contribution~\cite{NolteSupportingSafeDecision2020}.
In this contribution, we detailed how this influence manifests itself in the relation between scene elements and required skills in the skill graphs.
It would be interesting to evaluate the influence of other aspects of the ODD on the required vehicle skills.
The organization of this relational knowledge into an ontology also provides the opportunity of adding additional information such as monitoring metrics or monitoring requirements to this ontology as indicated in \cite{Nolteskillabilitybaseddevelopment2017}.

The ontology used in the presented approach for the automatic generation of skill graphs was adapted from a scene ontology for automatic scene generation.
Thus, at least two possible applications for an ontological representation of domain knowledge have been presented.
Additional applications in environment perception or scene understanding are evident possibilities.
As domain knowledge is required at several points during automated vehicle development, a single domain knowledge representation for all possible applications could be useful to limit inconsistencies during development.

%% file: sections/acknowledgment.tex

We would like to thank Marcus Nolte for the valuable discussions during conceptualization.